\documentclass[%
 reprint,
superscriptaddress,
showpacs,
 amsmath,amssymb,
 aps
]{revtex4-1}
\usepackage{times}
\usepackage{latexsym}

\usepackage{url}

\usepackage{graphicx}
\usepackage{dcolumn}
\usepackage{bm}
\usepackage{color}
\usepackage{here}
\newcommand{\tabitem}{~~\llap{\textbullet}~~}

\usepackage{url,listings}

\begin{document}
\title{Extractive Summarization via Weighted Dissimilarity and Importance Aligned Key Iterative Algorithm}
\author {Ryohto Sawada}

\begin{abstract}
  We present importance aligned key iterative algorithm for extractive summarization that is faster than conventional algorithms keeping its accuracy. The computational complexity of our algorithm is O($SNlogN$) to summarize original $N$ sentences into final $S$ sentences. Our algorithm maximizes the weighted dissimilarity defined by the product of importance and cosine dissimilarity so that the summary represents the document and at the same time the sentences of the summary are not similar to each other. The weighted dissimilarity is heuristically maximized by iterative greedy search and binary search to the sentences ordered by importance. We finally show a benchmark score based on summarization of customer reviews of products, which highlights the quality of our algorithm comparable to human and existing algorithms. 
\end{abstract}

\maketitle

\section{Introduction}
Automatic summarization of the document is in a great demand along with rapid growth of world wide web. There are two types of the automatic summarization; extractive summarization and abstractive summarization. Extractive summarization just selects important sentences from document. On the other hand, abstractive summarization revises the summary to compress redundancy and fill the lack of information \cite{radev1998, hongyan2000, nenkova2012,eisenstein2018}.

There are pros and cons for both extractive summarization and abstractive summarization. Extractive summarization requires relatively little domain knowledge. For example, one of the classical extractive summarization selects the sentences in the order of importance evaluated by counting frequency of the word \cite{fresch1948, luhn1958, luhn1957, sparch1972}. The counting does not require grammatical knowledge of the language and thus it is easy to expand to other languages. On the other hand, abstractive summarization is generally advantageous in accuracy \cite{ganesan2010, nayeem2018}. However, most of abstractive summarization requires additional knowledge of the language such as grammatical knowledge \cite{ganesan2010} and pre-trained word2vec embeddings \cite{mikolov2013, nayeem2018}. Therefore, abstractive summarization is generally more costly than extractive summarization. In this paper, we aim to present fast, accurate and multilingual extractive summarization algorithm due to following reasons; the amount of data on the internet is increasing by 2.5 quintillion bytes every day \cite{ibm2017} and there are about 200 languages on the internet \cite{w3tech2018}. 

Most of extractive summarization algorithms use the similarity between the sentences in the document. A common approach to characterize the sentences is vectorization that represents sentence $s$ as
\begin{eqnarray}
\label{eq:tfidf}
\vec{s} = \sum_{w \in s} c_{w} \vec{e}_{w},
\end{eqnarray}
where $w$ is the word in the sentence, $c_{w}$ is the weighting (importance) of the word and $\vec{e}_{w}$ is the unit vector satisfying $\vec{e}_{w} \cdot \vec{e}_{w'} = \delta_{ww'}$. 
The weighting of the word is usually calculated by term frequency and inverse document frequency (TFIDF) \cite{beel2016}.
TFIDF gives high weighting to the word that appears only in certain sentences \cite{fresch1948, luhn1958, luhn1957, sparch1972}. TDIDF helps to delete meaningless word like "the". The similarity between sentences can be evaluated by cosine product of the vectors.

There are various kinds of extraction algorithms using the similarities. A typical extraction algorithm is to maximize the minimum similarity between the sentences in the summary and the sentences in the document \cite{carbonell1998,radev2000, radev2004}. Graph theory also has high affinity for the extractive summarization by assuming the similarity as edge weight. For example, TextRank algorithm uses PageRank algorithm \cite{brin1998} to calculate importance of the sentences \cite{erkan2004, mihalcea2004}. Recently, Muhammad reported that an algorithm that uses the number of bicliques as the importance has as high accuracy as abstractive summarization \cite{muhammad2016}. 

However, these algorithms have problems with computational complexity and duplication. The ideal summarization should convey as much information as possible without duplication. In other words, the summary sentence should be similar to the sentence in the document, while the summary sentences should not be similar each other. However, computational complexity of the calculation of all the similarities is O($N^{2}$) for $N$ sentences. Furthermore, the maximization of the similarity between the summary and document is a vertex cover problem, typical example of an NP hard optimization problem. For the algorithms described in the previous paragraph, the computational complexities of TextRank algorithm and the biclique algorithm are O($N^{3}$) and O($N^{2}$), respectively. Additionally, the biclique algorithm gives up avoiding duplication and thus it is vulnerable to the duplication of the sentences [see Table.\ \ref{tab:sample} - \ref{tab:time} for detail]. 

In this paper, we solved the problem by importance aligned key iterative algorithm (IMAKITA). IMAKITA maximizes the weighted dissimilarity defined by the product of the importance and the dissimilarity so that the summary represents the document and at the same time the sentences of the summary are not similar to each other. 
The weighted dissimilarity can be heuristically maximized by iterative greedy search and binary search to the sentences ordered by importance. The computational complexity to summarize $N$ sentences to $S$ sentences is O($SN\log N$) in total. Furthermore, 
the benchmark score of IMAKITA is equivalent to humans in the summarization of customer reviews. Our algorithm will contribute to wide range of application of natural language processing. 

\section{Methods}

\begin{figure*}
  \begin{center}
    \includegraphics[width=16cm, bb = 0 0 1172 406 ]{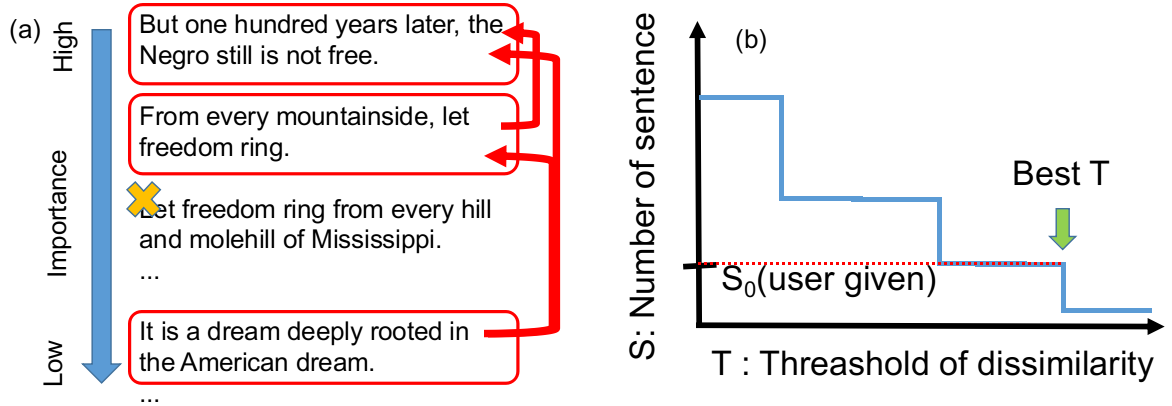}
    \caption{Schematics of the iterative greedy search (a) and outline of the function of the total number of the adopted sentences (b) .}
    \label{fig:greed}
  \end{center}
\end{figure*}


The weighted dissimilarity is calculated by the vectorized sentence defined by 
\begin{eqnarray}
\label{eq:vec}
\vec{s} = \frac{1}{N(s)} \sum_{w \in s} N_{w}\vec{e}_{w},
\end{eqnarray}
where $N(s)$ is the number of the word of the sentence $s$ and $N_{w}$ is the number of occurrences of the word $w$ in the document. As well as existing algorithms \cite{ganesan2010,muhammad2016}, we removed stopwords such as "is", "by", "the" from the sentences because they usually do not convey meaningful information. We used natural language toolkit for morphological analysis \cite{nltk2009} and library in \cite{chakki2018} for stopwords. This vectorization represents not only meaning of the sentence but also importance of the sentence. The importance of the sentence is defined by L2-norm of the vector $|\vec{s}^{2}|$. It means short sentences including the words frequently used in the document is important. This idea originates from well-known strategies for technical writing and presentation "Key phrase should be continuously used and message should be short" \cite{minto2008}. 
The weighted dissimilarity between sentence $s_{1}$ and $s_{2}$ is defined by
\begin{eqnarray}
\label{eq:dist}
D(\vec{s}_{1}, \vec{s}_{2}) = |\vec{s}_{1} | |\vec{s}_{2}| \sin \theta_{12}, 
\end{eqnarray}
where $\theta_{12}$ is angle between $\vec{s}_{1}$ and $\vec{s}_{2}$. The weighted dissimilarity becomes high only when both $s_{1}$ and $s_{2}$ are important and they have few words in common. Therefore, if a group of the sentences has high weighted dissimilarity, the group would carry important information without duplication. Unlike the existing algorithms, the maximization of the weighted dissimilarity can meet these two requirements at once.

The strict maximization of the weighted dissimilarity requires $O(_{N}C_{S})$ computational cost to summarize $N$ sentences to $S$ sentences. IMAKITA heuristically maximize the weighted dissimilarity by iterative greedy search and binary search to the sentences ordered by importance. Figure.\ \ref{fig:greed} (a) shows the schematics of iterative greedy search to find the group of the sentences whose dissimilarity is higher than certain value $T$. The sentences are evaluated in descending order of importance. When the minimum dissimilarity between the sentence and already accepted sentences is lower than $T$, the sentence will not be adopted. In this algorithm, the total number of the adopted sentences $S'(T)$ is the function of $T$ which monotonically decreases from $N$ to 1 with increase of $T$ [Figure.\ \ref{fig:greed} (b)]. The domain range of $T$ is $[0:|\vec{s}_{1}|^{2}]$ where $s_{1}$ is the most important sentence. When $S$ is given by user, the most appropriate $T$ is the largest T that satisfies $S'(T)=S$. This can be obtained by binary search because $S'(T)$ is monotonic function \cite{sedgewick2011}.
The binary search requires only whether $S'(T)$ is greater than or equal to $S$. Therefore, greedy iterative search be aborted when the number of adopted sentences reaches $S$. Furthermore, it is also possible to speed up the greedy iterative search by storing the weighted dissimilarity once calculated 
. The computational cost of the greedy search and binary search are $O(SN)$ and $O(\log N)$, respectively. Therefore, total computational complexity is $O(SN\log N)$. 

\section{Result}
Table\ \ref{tab:sample} shows the result of summarization of Martin Luther King Jr's speech \cite{king1963} using IMAKITA and existing algorithms.  We used summa library \cite{summa2018} for TextRank. Summary using biclique algorithm has duplication of content. Also, TextRank extracts unnecessarily long sentences because long sentences usually have something in common with other sentences. On the other hand, IMAKITA effectively summarizes the speech without duplication.

\begin{table*}[htb]
  \begin{tabular}{|l|l|}
    \hline
    \begin{tabular}{l}
      IMAKITA
    \end{tabular}
    & 
    \begin{tabular}{l}
    \tabitem   But one hundred years later, the Negro still is not free.\\
    \tabitem   And they have come to realize that their freedom is inextricably bound to \\
    our freedom.\\
    \tabitem   We cannot be satisfied as long as a Negro in Mississippi cannot vote and \\
    a Negro in New York believes he has nothing for which to vote.\\
    \tabitem   It is a dream deeply rooted in the American dream.\\
    \tabitem   From every mountainside, let freedom ring.\\
    \end{tabular}
    \\
    \hline
    \begin{tabular}{l}
      TextRank 
    \end{tabular}
    & 
    \begin{tabular}{l}
      \tabitem With this faith, we will be able to work together, to pray together, \\
      to struggle together, to go to jail together, to stand up for freedom\\
      together, knowing that we will be free one day.\\
      \tabitem And when this happens, when we allow freedom ring, when we let it ring\\
      from every village and every hamlet, from every state and every city,\\
      we will be able to speed up that day when all of God's children, black\\
      men and white men, Jews and Gentiles, Protestants and Catholics, will \\
      be able to join hands and sing in the words of the old Negro spiritual.
    \end{tabular}
    \\
    \hline
    \begin{tabular}{l}
      Biclique
    \end{tabular}
    & 
    \begin{tabular}{l}
      \tabitem But one hundred years later, the Negro still is not free.\\
      \tabitem Let freedom ring from the curvaceous slopes of California.\\
      \tabitem Let freedom ring from Lookout Mountain of Tennessee.\\
      \tabitem Let freedom ring from every hill and molehill of Mississippi.\\
      \tabitem From every mountainside, let freedom ring.
    \end{tabular}
    \\
    \hline
  \end{tabular}
  \caption{Result of the summarization of speech delivered by Martin Luther King Jr.}
  \label{tab:sample}
\end{table*}

The accuracy of the summarization is usually benchmarked by ROUGE score \cite{lin2004}. ROUGE evaluates the similarity between sentences based on an n-gram co-occurrence. The accuracy of the summarization can be estimated by ROUGE score between generated summaries and human summaries. We used Opinosis dataset \cite{ganesan2010} for the test data. Opinosis dataset provides user reviews on 51 different topics and summarization of the review by 5 different human workers. Original Opinosis paper also provides ROUGE scores between different human summaries and it is an important criterion for the accuracy of the summarization. Table\ \ref{tab:comp} shows the comparison of ROUGE scores between existing algorithms (human \cite{ganesan2010}, biclique \cite{muhammad2016}, sentence removal based on divergence and similarity (SR$_{DIV}$, SR$_{SIM}$) \cite{bonzanini2013}, TextRank \cite{mihalcea2004}, MEAD, Opinosis \cite{ganesan2010} and ParaFuse  \cite{nayeem2018}) and IMAKITA. The number of the summary sentences is set to 2. One can see IMAKITA is as accurate as humans and existing methods. 
\begin{table}[htb]
  \begin{tabular}{|l|l|l|l|}
    \hline
    Algorithm & R-1 & R-2 \\ 
    \hline
    Human & 30.88 & 10.69 \\ 
    \hline
    IMAKITA & 32.34 & 9.05 \\ 
    \ \ without stopwords  & 25.3 & 5.65 \\ 
    \ \ only importance & 29.65 & 6.5 \\ 
    \hline
    Biclique & 32.6 & 8.4 \\ 
    \hline
    SR$_{DIV}$  & 15.64 & 2.88 \\ 
    \hline
    SR$_{SIM}$ & 24.38 & 6.23 \\ 
    \hline
    TextRank & 27.56 & 6.12 \\ 
    \hline
    MEAD  & 15.15 & 3.08 \\ 
    \hline
    Opinosis  & 32.7 & 9.98 \\ 
    \hline
    ParaFuse & 33.86 & 9.74 \\ 
    \hline
  \end{tabular}
  \caption{Comparison of ROUGE scores between IMAKITA and existing algorithms.}
  \label{tab:comp}
\end{table}

\begin{table}[htb]
  \begin{tabular}{|l|l|l|l|}
    \hline
    Algorithm & AlphaZero & Alice & Carol \\
    \hline
    IMAKITA & 0.12 sec& 0.9 sec & 1.74 sec\\
    \hline
    Biclique & 0.18 sec& 1.35 sec & 3.71 sec\\
    \hline
    TextRank & 0.06 sec& 19 sec& 106 sec \\
   \hline
  \end{tabular}
  \caption{Comparison of calculation time between IMAKITA and existing algorithms.}
  \label{tab:time}
\end{table}

We also evaluated ROUGE score of IMAKITA without stopwords and the maximization of the weighted dissimilarity, respectively. Table\ \ref{tab:comp} shows both of them contribute to increase the accuracy of the summary. Surprisingly, IMAKITA without stopwords has F1-score as high as SR$_{SIM}$ algorithm despite having no domain knowledge in English except words are separated by space and sentences are separated by period, probably because the effect of the meaningless words are canceled out inside the iterative greedy search because most of sentences have these stopwords. The result implies IMAKITA is easy to apply to other languages.

Table\ \ref{tab:time} shows the comparison of calculation time between IMAKITA and existing algorithms. We used Intel Corei7-4710MQ for the calculation and set the number of the summary sentences to 10. We used AlphaZero paper \cite{silver2018} (2523 words, 126 sentences) , Alice's Adventures in Wonderland \cite{carroll1865} (15192 words, 501 sentences) and A Christmas Carol for the benchmark \cite{dickens1843} (29140 words, 1420 sentences). One can see that the calculation time of IMAKITA is approximately linear unlike existing algorithms.

\section{Conclusion}
We presented importance aligned key iterative algorithm for extractive summarization that is faster than conventional algorithms keeping its accuracy. The algorithm maximizes the weighted dissimilarity so that the summary represents the document and at the same time the sentences of the summary are not similar to each other. The weighted dissimilarity is heuristically maximized by iterative greedy search and binary search. We demonstrated that ROUGE score of IMAKITA is comparable to human and existing algorithms and computation time is shorter than these algorithms. We also analyzed the contributuion of the weighted dissimilarity and stopwords and showed IMAKITA works by little domain knowledge. Our algorithm will contribute to wide range of application of natural language processing.

\section{Appendix}

\subsection{Source code}
We provide the source code of IMAKITA for Chrome Extension on github \url{https://github.com/qhapaq-49/imakita}.

\subsection{Pseudo code}
\begin{lstlisting}[tabsize=3, caption=Pseudo code of IMAKITA search.,label=pcode]

for sentence in document:
	for word in sentence:
		document[word] += 1

for sentence in document:
	for word in sentence:
		vector[sentence][word] += document[word] / word_count[sentence]
		importance[sentense] += document[sentence][word] ** 2
sort(importance)
output = binary_search(importance, number_of_summary)

Function binary_search(importance, number_of_summary):
	output = None
	pos = max(importance) ** 2 / 2
	step = pos / 2
	while step > epsilon:
		if check(importance, number_of_summary, pos, output):
			pos += step
		else:
			pos -= step
		step = step / 2
	return output
                
Function check(importance, number_of_summary, threshold, output):
	// greedy search
	result = None
	itr = 0
	for im in importance:
		for res in result:
			if cross(im, res) < threshold:
				continue
		result[itr] = im
		itr += 1
		if itr == number_of_summary:
			output=result
			return True
	return False                        
        
Function cross(sentence, sentence2):
	norm1 = product(sentence1, sentence1)
	norm2 = product(sentence2, sentence2)
	cdot = product(sentence1, sentence2)
	cos12 = cdot / sqrt(norm1*norm2)
	return sqrt(norm1 * norm  * (1 - cos12 * cos12))

Function product(sentence1, sentence2):
	for word1 in sentence1:
		for word2 in sentence2:
			if word1 == word2:
				output += document[word]  / (word_count[sentence1] * word_count[sentence2])
	return output

\end{lstlisting}

\bibliography{swdbib}
\bibliographystyle{junsrt}
\end{document}